\pdfoutput=1

\documentclass[11pt]{article}

\usepackage[]{ACL2023}

\usepackage{times}
\usepackage{latexsym}

\usepackage[T1]{fontenc}
\usepackage{longtable}
\usepackage[utf8]{inputenc} 
\usepackage[T1]{fontenc}    
\usepackage{hyperref}       
\usepackage{url}            
\usepackage{booktabs}       
\usepackage{amsfonts}       
\usepackage{nicefrac}       
\usepackage{microtype}      
\usepackage{xcolor}         
\usepackage{colortbl}
\usepackage{multirow}
\usepackage{soul} 
\usepackage{booktabs}
\usepackage{graphicx}
\usepackage{subcaption}
\usepackage{graphics}
\usepackage{xspace}
\usepackage{amsmath}
\usepackage{amssymb}  
\usepackage{adjustbox}
\usepackage{makecell}
\usepackage{pifont}
\usepackage{tabularx}
\usepackage{comment}

\usepackage{fontawesome} 

\newcommand{\ours}{{\textsc{Socialite-llama}}\xspace}
\newcommand{\ourdata}{{\textsc{SocialiteInstructions}}\xspace}

\newcommand{\llama}{{\textsc{Llama2}}\xspace}
\newcommand{\deberta}{{\textsc{DeBERTa}}\xspace}

\newcommand{\llm}{\textsc{LLM}}
\newcommand{\gptfour}{\textsc{GPT-4}\xspace}
\newcommand{\socket}{\textsc{SocKET}\xspace}
\newcommand{\sbf}{\textsc{Social Bias Frames}\xspace}

\newcommand{\humcls}{\textsc{Humor}\xspace}
\newcommand{\humrat}{\textsc{HumourRating}\xspace}
\newcommand{\flute}{\textsc{FLUTE}\xspace}
\newcommand{\offense}{\textsc{Offensive}\xspace}
\newcommand{\sexyn}{\textsc{Sexist}\xspace}
\newcommand{\intent}{\textsc{IntentToOffend}\xspace}
\newcommand{\biasedimp}{\textsc{BiasedImplication}\xspace}
\newcommand{\emotion}{\textsc{Emotion}\xspace}
\newcommand{\Sentiment}{\textsc{Sentiment}\xspace}
\newcommand{\sameside}{\textsc{SameSideStance}\xspace}
\newcommand{\valence}{\textsc{ValenceCLS}\xspace}
\newcommand{\arousal}{\textsc{ArousalCLS}\xspace}
\newcommand{\dominance}{\textsc{DominanceCLS}\xspace}
\newcommand{\hyperbole}{\textsc{Hyperbole}\xspace}
\newcommand{\neutralisebias}{\textsc{SubjectiveBias}\xspace}

\newcommand{\empathyself}{\textsc{EmpathySelfRated}\xspace}
\newcommand{\distressself}{\textsc{DistressSelfRated}\xspace}

\newcommand{\empathy}{\textsc{EmpathyExplorations}\xspace}
\newcommand{\politenesshayati}{\textsc{PolitenessHayati}\xspace}
\newcommand{\intimacy}{\textsc{Intimacy}\xspace}

\newcommand{\hate}{\textsc{HateSpeech}\xspace}
\newcommand{\irony}{\textsc{Irony}\xspace}
\newcommand{\stanfordpoliteness}{\textsc{PolitenessStanford}\xspace}
\newcommand{\optimism}{\textsc{Optimism}\xspace}
\newcommand{\complaints}{\textsc{Complaints}\xspace}
\newcommand{\kialo}{\textsc{AgreeDisagree}\xspace}

\usepackage{todonotes}

\title{\textsc{\ours}: An Instruction-Tuned Model\\ for Social Scientific Tasks}

\author{Gourab Dey$^1$\thanks{\, Equal contribution}\ , Adithya V Ganesan$^1$$^*$, Yash Kumar Lal$^1$$^*$,\\ \textbf{ Manal Shah$^2$, Shreyashee Sinha$^3$, Matthew Matero$^4$} \\ \textbf{Salvatore Giorgi$^{5,6}$, Vivek Kulkarni$^7$, H. Andrew Schwartz$^1$} \\
  $^1$Stony Brook University,
  $^2$JP Morgan Chase,
  $^3$Bloomberg \\
  $^4$Preferabli,
  $^5$University of Pennsylvania,
  $^6$National Institute of Health, 
  $^7$Grammarly \\
  $^*$\texttt{\{gdey,avirinchipur,ylal\}@cs.stonybrook.edu}\\ \\}

\begin{document}
\maketitle

\begin{abstract}
Social science NLP tasks, such as emotion or humor detection, are required to capture the semantics along with the implicit pragmatics from text, often with limited amounts of training data. 
Instruction tuning has been shown to improve the many capabilities of large language models (\llm s) such as commonsense reasoning, reading comprehension, and computer programming.
However, little is known about the effectiveness of instruction tuning on the social domain where implicit pragmatic cues are often needed to be captured.
We explore the use of instruction tuning for social science NLP tasks and introduce \ours --- an open-source, instruction-tuned \llama.
On a suite of 20 social science tasks, \ours improves upon the performance of \llama as well as matches or improves upon the performance of a state-of-the-art, multi-task finetuned model on a majority of them.
Further, \ours also leads to improvement on 5 out of 6 related social tasks as compared to \llama, suggesting instruction tuning can lead to generalized social understanding.
All resources including our code, model and dataset can be found through ~\href{https://bit.ly/socialitellama}{\texttt{bit.ly/socialitellama}}.


\end{abstract}

\section{Introduction}

Instruction finetuned large language models (\llm s) have demonstrated impressive performance on many standard NLP tasks \cite{wei2022finetuned, chung2022scaling}, but these models tuned on non-social tasks seem to have poor social pragmatics \cite{ziems2023can, choi2023llms, havaldar-etal-2023-multilingual, v-ganesan-etal-2023-systematic}.
The fact that instruction tuning has been successful with limited amounts of data \cite{gupta2023instruction} and that instruction tuned models have the ability to generalize to new tasks in both few- and zero-shot settings \cite{wei2022finetuned}, suggests that \llm s could become more socially capable by instruction tuning them on a wide variety of social NLP tasks. 




Here, we introduce \ours, a \llama 7B-based \llm ~\cite{touvron2023llama} instruction tuned on a suite of social scientific classification tasks spanning 5 broad categories for which we hand-craft instructions. 
We evaluate its performance on both zero- and few-shot settings on seen and related social tasks, demonstrating that our model significantly outperforms prior open models.
Our results support prior research highlighting the effectiveness of instruction tuning when applying \llm s to a new domain.

Social and psychological factors have been shown important and beneficial to model in past interdisciplinary NLP studies~\cite{lynn-etal-2017-human, flek-2020-returning, hovy-yang-2021-importance}. 
Modeling human factors and social context can not only improve performance on primarily non-social NLP tasks ~\cite{lynn-etal-2017-human, flek-2020-returning, hovy-yang-2021-importance}, but can also prove to be beneficial for a number of psychological and social scientific tasks ~\cite{garten_incorporating_2019, matero2021melt}.
However, a major limitation of prior models has been that they have been task-specific and do not generalize well to new tasks. 
Many such models also are trained from scratch and typically need an extensive amount of human-annotated training data. 
With the development of \llm s that exhibit the capability to learn from instructions, we posit that this new capability can be leveraged to address these two major limitations.  



Our \textbf{contributions} include: (1) we develop and systematically evaluate \ours -- an instruction-tuned language model for social science tasks -- across 20 seen and 6 related social scientific tasks, 
(2) we show \ours consistently improved over \llama in all seen tasks and the improvement generalized to 5 of 6 related tasks; In fact, it matched the performance of a state-of-the-art multi-task tuned \deberta on a majority of seen tasks, 
(3) we suggest that the benefits of few-shot examples (over zero-shot) become negligible on tasks seen during instruction tuning as opposed to related tasks where few-shot still provided a benefit, 
and (4) we release \ours\footnote{Model available on \href{https://huggingface.co/hlab/SocialiteLlama}{huggingface models}} as well as its instructions corpora, \ourdata\footnote{Dataset available on \href{https://huggingface.co/datasets/hlab/SocialiteInstructions}{huggingface datasets}}, as open-source resources for the community.

\section{Related Work}


Language is inherently social --- the underlying meaning is constructed through social interactions \cite{wittgenstein1953philosophical,  Clark1992AskingQA, hovy-spruit-2016-social}.
Understanding communication requires reasoning about the social implications drawn from that message \cite{halliday2004introduction}.
Prior work has sought to build language models for social scientific tasks, which we discuss next.

\paragraph*{LMs for Social Scientific NLP}
Social science NLP models are usually built by fine-tuning \cite{sap-etal-2020-social, matero2019suicide, v-ganesan-etal-2021-empirical} for specific tasks or pretraining language models on the the domain of language that captures the social factors. 
BERTweet \cite{nguyen-etal-2020-bertweet}, the first public large-scale pre-trained language model on English tweets, was trained using the BERT architecture with the RoBERTa pre-training procedure \cite{liu2019roberta} for this purpose. 
\citet{delucia-etal-2022-bernice} built Bernice, a multilingual model for social science NLP tasks using the RoBERTa architecture and pretrained from scratch on 2.5 billion tweets.
It outperforms a variety of models adapted to social media data as well as strong multilingual baselines. 
Despite the strength of such existing task-specific models, there exists no base instruction-following \llm~tailored to the domain of social science tasks.


\paragraph*{Instruction-Tuning}
Instruction tuning, in general, refers to the practice of finetuning pre-trained language models to better understand and respond to a wide variety of human requests that are expressed in natural language \cite{wei2022finetuned, mishra-etal-2022-cross}.
The success of instruction tuning requires two key components: 1) a powerful pre-trained language model like \llama \cite{touvron2023llama}, and 2) an instruction dataset that is diverse and representative enough to adapt the \llm~to potential downstream usage.
However, existing instruction-tuning datasets \cite{flan-data, sanh2022multitask, wang-etal-2023-self-instruct, wang-etal-2022-super} are general-purpose and do not contain a significant amount of social science tasks.
Consequently, models built using this data have limited usability for social scientific tasks \cite{choi2023llms, ziems2023can}.
To address this gap, we curate an instruction-tuning dataset aimed at modeling social scientific knowledge and use it to train our model.

\section{Datasets and Experiments}

We use a diverse collection of English social science NLP tasks to create \ourdata by (re-)framing each task into a binary or multi-class classification problem.
For each task, we hand-craft instructions for the model to follow. These instructions were composed with task descriptive which included examples in some cases (e.g. \biasedimp from ~\autoref{tab:prompts-offense}), followed by the classification label choices it has to pick from.   
The instructions and the dataset sizes for all the tasks have been tabulated in \autoref{sec:appendix}. 


\subsection{Training Tasks}

We draw on datasets mentioned in \citet{choi2023llms} for compiling our instruction following dataset. 
Our train set spans 20 different datasets across 5 broad categories of tasks: Humor, Offensiveness, Sentiment and Emotion, Trustworthiness, and other social factors.
To ensure broader applicability, we frame the non-classification datasets as classification tasks.
We refer to these as seen tasks going forward.

\textbf{Humor}
We use SemEval 2021 Task 7 data \cite{meaney-etal-2021-semeval} to capture humor.
We use the binary humor detection task as is (\humcls).
Additionally, we convert the humor rating task into a binary classification problem collapsing ground truth labels higher than 3 into high humor, and the others into low humor (\humrat).
 
\textbf{Offensiveness} 
We use \sbf \cite{sap-etal-2020-social} as a benchmark to detect offense and bias directed towards people and groups.
\sbf comprises of 4 binary classification tasks - \offense to discern whether a given text exhibits rudeness, disrespect, or toxicity, \sexyn to determine if a text contains lewd or sexual references which can be considered as offensive, \intent to capture whether the perceived motivation of the author was to indeed offend, and \biasedimp to identify any forms of prejudice or group-based discrimination within the text.

\textbf{Sentiment and Emotion}
For emotion classification, we use SemEval 2018 Task 1 data \cite{mohammad-etal-2018-semeval}.
\emotion involves classifying a tweet into anger, joy, optimism or sadness displayed by the author.
We also include Emobank \cite{buechel-hahn-2017-emobank} to infer the valence, arousal and dominance levels that a text would invoke in a reader. 
We convert these regression tasks to binary classification - by transforming a score greater than 4 to `high' and less than 4 to `low', for each of the three datasets respectively (\valence, \arousal, \dominance).
We use SemEval 2017 Task 4 data for sentiment classification \cite{rosenthal-etal-2017-semeval} to identify the overall sentiment of a text as positive, negative or neutral.
Further, we also include \sameside \cite{korner-etal-2021-classifying}, the task of classifying whether two different texts are on the same side of an argument.

\textbf{Trustworthiness}
We use \hyperbole \cite{zhang-wan-2022-mover} to detect hyperbolic language present in a text, and \neutralisebias \cite{Subjectivebias}, to detect which sentence exhibits subjective bias among two pieces of text.

\textbf{Other Social Factors}
We also include other discursive and rhetorical type tasks --- empathy scored on multi-item scales (\empathy) \cite{sharma-etal-2020-computational}, self-rated empathy (\empathyself) \cite{buechel-etal-2018-modeling}, distress (\distressself)~\cite{buechel-etal-2018-modeling}, figurative speech detection (\flute)~\cite{chakrabarty-etal-2022-flute}, politeness (\politenesshayati)~\cite{hayati-etal-2021-bert} and intimacy (\intimacy)~\cite{pei-jurgens-2020-quantifying}.

\subsection{Evaluation Tasks}

For evaluation, we use the same twenty seen datasets as well as six datasets for related social tasks.
The process we used to choose the ‘seen tasks’ for instruction tuning was towards the goal of training on a very broad category of tasks and consequently, any task would be related to these fundamental categories seen during training. Hence for evaluation we choose six tasks, half of which are more directly related to seen tasks than the other. The datasets used for evaluation contained no data overlap with the training tasks' data.
Among the related tasks' datasets, we first select three datasets directly related to our training tasks --- \hate~\cite{basile-etal-2019-semeval} to detect hate speech in a tweet, \stanfordpoliteness \cite{fu-etal-2020-facilitating} to judge whether a text is polite or impolite, and \kialo \cite{varadarajan-etal-2022-detecting} to classify whether two texts agree, disagree or N/A on a particular topic.
To strengthen our evaluation, we also evaluate on three other tasks which are less related to the seen set --- \irony~\cite{van-hee-etal-2018-semeval} to classify if a tweet is ironic, \optimism~\cite{ruan-etal-2016-finding} to categorize a text as optimistic, pessimistic or neutral, and \complaints~\cite{preotiuc-pietro-etal-2019-automatically} to judge if a text is a complaint or not.
We refer to these six as related social tasks.

\subsection{Task Selection Criterion}
The above tasks were chosen from \socket (a) to be representative of each social category and (b) if the document lengths weren’t too long, i.e., utterance level tasks~\cite{ziems2023can}.

Our initial experiments suggested that skew in sample sizes from different tasks affected the overall performance. Particularly, using all 35k examples from \sbf for all the four tasks limited the performance on \politenesshayati and \sameside. Hence, we picked datasets from \socket based on 2 additional factors: recency and minimum number of 8k examples. This is why we chose \citet{sap-etal-2020-social} over other toxicity datasets (further down sampled 8k examples for each task), and \citet{rosenthal-etal-2017-semeval} over \citet{socher-etal-2013-recursive} for sentiment task.

\begin{table*}[!t]
\centering
\begin{tabular}[width=\columnwidth]{l|cc|cc}
\hline
\textbf{\textsc{Task}} & \multicolumn{2}{c|}{\begin{tabular}{@{}c@{}}\textbf{\textsc{Socialite-}} \\ \textbf{\textsc{Llama}} \end{tabular}} &  \multicolumn{2}{c}{\textbf{\llama}} \\
{} & Few-shot & Zero-shot & Few-shot & Zero-shot \\
\hline \hline
\multicolumn{5}{c}{\textsc{Seen Tasks}} \\
\hline
\Sentiment        & 72.69 & \textbf{73.43}  & 46.40 & 18.28 \\
\emotion          & 79.68 & \textbf{81.08}  & 66.84 & 23.92 \\
\valence          & 82.34 & \textbf{83.84}  & 44.52 & 31.59 \\
\arousal          & 60.69 & \textbf{64.22}  & 52.69 & 34.42 \\
\dominance        & 60.88 & \textbf{66.63}  & 43.05 & 19.33 \\
\empathy   & 30.10 & \textbf{56.99} & 29.43 & 5.89 \\
\empathyself      & 63.59 & \textbf{63.97}  & 58.99 & 8.75 \\
\distressself & 68.45 & \textbf{70.96} & 42.36 & 9.80 \\
\flute            & 80.77 & \textbf{96.27} & 55.13 & 4.81 \\
\hyperbole        & 65.62 & \textbf{69.23} & 59.20 & 25.43 \\
\sameside         & \textbf{88.19} & \textbf{88.19} & 82.10 & 25.00 \\
\humcls           & 90.48 & \textbf{95.29} & 44.44 & 38.8 \\
\humrat           & \textbf{52.81} & 48.66 & 43.51 & 34.84 \\
\politenesshayati & \textbf{89.36} & 84.54 & 83.58 & 24.34 \\
\intimacy         & 26.25 & \textbf{33.21} & 13.24 & 3.17 \\
\neutralisebias   & 52.04 & \textbf{87.13} & 51.18 & 40.98 \\
\offense          & 80.47 & \textbf{83.11} & 71.78 & 35.76 \\
\sexyn            & 72.61 & \textbf{73.85} & 52.81 & 6.25 \\
\intent           & 74.17 & \textbf{75.65} & 32.09 & 21.34 \\
\biasedimp        & 79.66 & \textbf{85.72} & 56.41 & 26.16 \\
\hline \hline
\multicolumn{5}{c}{\textsc{Related Tasks}} \\ 
\hline
\hate       & 42.13 & 39.63 & \textbf{68.66} & 31.34 \\
\irony      & 42.99 & \textbf{59.04} & 47.11 & 32.24 \\
\stanfordpoliteness &  63.22 & \textbf{64.89} & 58.64 & 1.2 \\
\optimism   & \textbf{62.24} & 36.96 & 56.61 & 44.49 \\
\complaints & 79.35 & \textbf{85.37} & 58.23 & 22.8 \\
\kialo      & \textbf{39.54} & 19.3  & 28.1  & 10.01 \\
\hline

\end{tabular}
\caption{Evaluation of \ours and \llama on 20 seen tasks and 6 unseen tasks using macro-F1 scores. We report few-shot performance with $k=5$ for \ours on seen tasks and $k=15$ on unseen tasks. For \llama, we use $k=15$ for all tasks since it is not trained for social scientific tasks. $k$ refers to the number of examples used in the few-shot setting. All differences between \ours and \llama are statistically significant $(p<0.001)$.
\textbf{Bold} indicates the performance with the best setting for that task.
}
\label{tab:instruction-tune-eval}
\end{table*}

\subsection{Training Procedure}

We transform each of the 20 training datasets using hand-crafted instructions. 
To reduce the skew caused by the disproportionate number of examples present in a few tasks, we limit the number of training examples from \offense, \sexyn, \intent, \biasedimp, \Sentiment and \neutralisebias to 8k and use the original number of training data points for the rest. This resulted in \textasciitilde 108k data points. 
We employ Low-Rank Adaptation \cite{lora}, with rank set to 8 applied on the key and query projection matrices reducing the number of trainable parameters to 4.1 million, and use AdamW \cite{adamw} with a learning rate of $\eta=1e-4$ and a batch size of $64$ for a maximum of $7$ epochs with early stopping.
We expand the context length to 3k using mixed precision training, gradient checkpointing and DeepSpeed Zero \cite{rajbhandari2020zero} Stage 2 which enables finetuning on 2 48GB A6000 GPUs in \textasciitilde 21 hours and train using Huggingface \cite{wolf-etal-2020-transformers}.
Training \ours on Lambda\footnote{\url{https://lambdalabs.com}} would cost \$33.60, making it very cost-effective.

\section{Results}



We analyze the effect of instruction tuning \llama on 20 seen and 6 related social scientific tasks and report macro F1 scores in \autoref{tab:instruction-tune-eval}.

\noindent \textbf{Instruction tuning improves social understanding.} We show that \ours outperforms \llama on all seen tasks in both zero- and few-shot setting.
In fact, \ours used in a zero-shot setting is even better than \llama in few-shot setting.
Surprisingly, we find that the benefit of few-shot learning over zero-shot is much less on seen tasks as opposed to the related tasks with the instruction tuned model.

Particularly, performance of our model is better for all tasks that are proximate to personal factors or how one feels, such as the affective dimensions, sentiment, emotion and empathy/distress. 
We also see consistent improvements over \llama on offensiveness related tasks.
Despite the tremendous general-purpose capabilities of \llama, it appears to lack social understanding from text, and can be improved using instruction tuning on social scientific tasks.

\begin{table}[!t]
\centering
\small
\setlength\tabcolsep{4.3pt}
\renewcommand{\arraystretch}{1.15}
\begin{tabular}{ l  |cc  }
\hline
\textbf{\textsc{Task}}  &\begin{tabular}{@{}c@{}}\textbf{\textsc{Socialite-}} \\ \textbf{\textsc{LLama}} \\
\end{tabular}  & \begin{tabular}{@{}c@{}} \textbf{\deberta} \\ \small{(multi-task FT)} \end{tabular} \\
\hline
\Sentiment  & \textbf{73.43}  & 69.00 \\
\emotion  & \textbf{81.08}  & 80.00 \\
\distressself & \textbf{70.96} & 65.00 \\
\empathyself & \textbf{63.97} & 59.00 \\
\hyperbole  & \textbf{69.23} & 69.00 \\
\sameside  & \textbf{88.19}  & 76.00 \\
\humcls  & \textbf{95.29} & 91.00 \\
\politenesshayati  & \textbf{84.54} & 89.00 \\
\intimacy  & 33.21 & 46.00 \\
\neutralisebias  & 87.13 & 96.00 \\
\offense  & \textbf{83.11} & 83.00 \\
\intent  & \textbf{75.65} & 74.00 \\
\sexyn  & 73.85 & 79.00 \\
\biasedimp  & 85.72 & 87.00 \\
\hline
\end{tabular}
\caption{
We compare \ours with a multi-task finetuned (FT) \deberta reported in \citet{choi2023llms}.
\ours matches or outperforms it on 10 out of 14 tasks that both models have been trained on, even though our training data size is much lower.
\textbf{Bold} macro F1 represents improvement or equivalent performance of \ours with \deberta (statistically significant results performed using bootstrapped resampling; $p<.05$ ).
Because \deberta is not generative, it is not possible to apply to tasks for which it was not finetuned. 
}
\label{tab:multi-task}
\end{table}

\noindent \textbf{Instruction tuning helps generalization.}
In related social science tasks, which are composed of new instructions and language samples, we find that \ours is better on 4 out of 6 under few-shot setting and 5 out of 6 overall. 
These 6 related tasks were on a varying degree of relatedness to the ones in the seen tasks.
\ours shows significant improvement in performance over \llama on the related social science tasks.
Through instruction tuning, our model generalizes social aspects of language to perform well on new related social tasks and task categories.

To assess the generalization emerging from the commonalities present between the task, we applied the instructions from the seen tasks' set on a related task and compared its performance with the related tasks' instruction (which was unseen to \ours). A non-author NLP expert suggested 4 seen tasks (\offense, \sexyn, \intent and \biasedimp) that they thought were very similar to the \hate task. Of those, instructions from \offense and \intent gave improvement over zero-shot \hate performance, while the other two led to worse performance.  This suggests it's possible to do better on zero-shot by using a related task prompt from instruction tuning but it can also hurt performance.


\noindent \textbf{\ours is state-of-the-art overall on \socket.}
We compare \ours with a state-of-the-art multi-task finetuned \deberta model presented in \citet{choi2023llms} for a subset of the seen tasks since their \deberta model can't be applied on the related social tasks. 
Despite being trained on significantly less data than \deberta model (2.1 million data points and 58 tasks as compared to 108k data points and 20 tasks), \autoref{tab:multi-task} shows that our model is better or equivalent in performance on 10 out 14 tasks. This result highlights a large reduction in the performance separation between \llm s and smaller task finetuned language models in Social Science tasks established by previous works~\cite{ziems2023can, choi2023llms}.    



\section{Conclusion}

We introduce \ours, an instruction tuned \llama 7B for social science applications. 
\ours is trained using \ourdata, a diverse collection of 20 social scientific tasks and tested on those plus an additional 6 related social tasks.
It consistently performs better than \llama on all the seen tasks and all but 1 related task, demonstrating its strong generalization abilities.
Instruction tuning on diverse social scientific tasks helps our model achieve state-of-the-art overall.
Our results provide valuable insights to improve the social understanding of \llm s.


\section*{Limitations}
The improvements brought through instruction tuning of \llama on social scientific tasks highlights the prevailing neglect of these tasks and the importance of doing this. 
However, there are some important limitations. 
For one, the tests for generalization to unseen tasks were limited to those that are semantically related to seen tasks (many tasks in \socket are related to each other).
Future work would need to evaluate this model and process for generalizing to more distant tasks as well as across other points in time. 

This work was informed by a number of works in the NLP and deep learning literature, was largely facilitated by various open sourced datasets~\cite{choi2023llms} and code~\cite{wolf-etal-2020-transformers}. 
However, the number of tasks in our work is not close to the number of tasks found in the prior instruction tuning literature~\cite{wang-etal-2022-super, sanh2022multitask}. 
This work would likely benefit more tasks within the realm of social science NLP. 

Owing to limited access to computational resources, we could only focus on classification tasks on the smallest available \llama model and Low-Rank Adaptation as training strategy. 
However, literature on scaling models~\cite{gpt3, wei2022emergent} suggests that these performances likely go up with increasing the model size, datasets, instructions, which would make it more capable for regression and generation tasks with increased input lengths. 
Inclusion of few-shot examples in the training set could also improve the few-shot generalization capability of the model, which could not be performed with the resources available.        

\section*{Ethics Statement}
Large language models (\llm s) like \gptfour~\cite{gpt4techreport} have been deployed to tens of millions of consumers in different forms \cite{Heaven2023ChatGPT, Hu2023ChatGPT}. 
While a number of works have already shown its limited social understanding~\cite{ziems2023can, choi2023llms, havaldar-etal-2023-multilingual, v-ganesan-etal-2023-systematic}, their rapid adoption and proliferation necessitates improving their social sensibility to make it reliable and useful. 
This work is aimed towards building a socially sensible language model for the practitioners of social science NLP research.

We hope our work improves NLP's benefit for social scientific pursuits and caution against the use of such models for non-scientific pursuits like targeting of individuals without their consent or awareness.
This work has been built for research and can be re-distributed as per the policy laid out by Meta for \llama\footnote{\url{https://ai.meta.com/llama/license/}}.
We strongly believe that these models have to be tested for failure modes and harmful biases, and should be further be adjusted through appropriate methods~\cite{ouyang2022training} before deploying it into the hands of users.

\section*{Acknowledgements}

We would like to thank the reviewers for their valuable feedback that helped us improving this paper. This work was supported in part by the Office of the Director of National Intelligence (ODNI), Intelligence Advanced Research Projects Activity (IARPA), via the HIATUS Program contract (585968), a grant from the CDC/NIOSH (U01 OH012476), and a grant from the NIH-NIAAA (R01 AA028032). YKL was supported in part by the Air Force Research Laboratory (AFRL), DARPA, for the KAIROS program under agreement number FA8750-19-2-1003 and in part by the NSF under the award IIS 2007290. The views and conclusions contained herein are those of the authors and should not be interpreted as necessarily representing the official policies, either expressed or implied, of ODNI, IARPA, any other government organization, or the U.S. Government. The U.S. Government is authorized to reproduce and distribute reprints for governmental purposes notwithstanding any copyright annotation therein.

\bibliographystyle{acl_natbib}
\bibliography{anthology,custom}

\begin{thebibliography}{57}
\expandafter\ifx\csname natexlab\endcsname\relax\def\natexlab#1{#1}\fi

\bibitem[{Basile et~al.(2019)Basile, Bosco, Fersini, Nozza, Patti, Rangel~Pardo, Rosso, and Sanguinetti}]{basile-etal-2019-semeval}
Valerio Basile, Cristina Bosco, Elisabetta Fersini, Debora Nozza, Viviana Patti, Francisco~Manuel Rangel~Pardo, Paolo Rosso, and Manuela Sanguinetti. 2019.
\newblock \href {https://doi.org/10.18653/v1/S19-2007} {{S}em{E}val-2019 task 5: Multilingual detection of hate speech against immigrants and women in {T}witter}.
\newblock In \emph{Proceedings of the 13th International Workshop on Semantic Evaluation}, pages 54--63, Minneapolis, Minnesota, USA. Association for Computational Linguistics.

\bibitem[{Brown et~al.(2020)Brown, Mann, Ryder, Subbiah, Kaplan, Dhariwal, Neelakantan, Shyam, Sastry, Askell, Agarwal, Herbert-Voss, Krueger, Henighan, Child, Ramesh, Ziegler, Wu, Winter, Hesse, Chen, Sigler, Litwin, Gray, Chess, Clark, Berner, McCandlish, Radford, Sutskever, and Amodei}]{gpt3}
Tom Brown, Benjamin Mann, Nick Ryder, Melanie Subbiah, Jared~D Kaplan, Prafulla Dhariwal, Arvind Neelakantan, Pranav Shyam, Girish Sastry, Amanda Askell, Sandhini Agarwal, Ariel Herbert-Voss, Gretchen Krueger, Tom Henighan, Rewon Child, Aditya Ramesh, Daniel Ziegler, Jeffrey Wu, Clemens Winter, Chris Hesse, Mark Chen, Eric Sigler, Mateusz Litwin, Scott Gray, Benjamin Chess, Jack Clark, Christopher Berner, Sam McCandlish, Alec Radford, Ilya Sutskever, and Dario Amodei. 2020.
\newblock \href {https://proceedings.neurips.cc/paper_files/paper/2020/file/1457c0d6bfcb4967418bfb8ac142f64a-Paper.pdf} {Language models are few-shot learners}.
\newblock In \emph{Advances in Neural Information Processing Systems}, volume~33, pages 1877--1901. Curran Associates, Inc.

\bibitem[{Buechel et~al.(2018)Buechel, Buffone, Slaff, Ungar, and Sedoc}]{buechel-etal-2018-modeling}
Sven Buechel, Anneke Buffone, Barry Slaff, Lyle Ungar, and Jo{\~a}o Sedoc. 2018.
\newblock \href {https://doi.org/10.18653/v1/D18-1507} {Modeling empathy and distress in reaction to news stories}.
\newblock In \emph{Proceedings of the 2018 Conference on Empirical Methods in Natural Language Processing}, pages 4758--4765, Brussels, Belgium. Association for Computational Linguistics.

\bibitem[{Buechel and Hahn(2017)}]{buechel-hahn-2017-emobank}
Sven Buechel and Udo Hahn. 2017.
\newblock \href {https://aclanthology.org/E17-2092} {{E}mo{B}ank: Studying the impact of annotation perspective and representation format on dimensional emotion analysis}.
\newblock In \emph{Proceedings of the 15th Conference of the {E}uropean Chapter of the Association for Computational Linguistics: Volume 2, Short Papers}, pages 578--585, Valencia, Spain. Association for Computational Linguistics.

\bibitem[{Chakrabarty et~al.(2022)Chakrabarty, Saakyan, Ghosh, and Muresan}]{chakrabarty-etal-2022-flute}
Tuhin Chakrabarty, Arkadiy Saakyan, Debanjan Ghosh, and Smaranda Muresan. 2022.
\newblock \href {https://doi.org/10.18653/v1/2022.emnlp-main.481} {{FLUTE}: Figurative language understanding through textual explanations}.
\newblock In \emph{Proceedings of the 2022 Conference on Empirical Methods in Natural Language Processing}, pages 7139--7159, Abu Dhabi, United Arab Emirates. Association for Computational Linguistics.

\bibitem[{Choi et~al.(2023)Choi, Pei, Kumar, Shu, and Jurgens}]{choi2023llms}
Minje Choi, Jiaxin Pei, Sagar Kumar, Chang Shu, and David Jurgens. 2023.
\newblock \href {https://doi.org/10.18653/v1/2023.emnlp-main.699} {Do {LLM}s understand social knowledge? evaluating the sociability of large language models with {S}oc{KET} benchmark}.
\newblock In \emph{Proceedings of the 2023 Conference on Empirical Methods in Natural Language Processing}, pages 11370--11403, Singapore. Association for Computational Linguistics.

\bibitem[{Chung et~al.(2022)Chung, Hou, Longpre, Zoph, Tay, Fedus, Li, Wang, Dehghani, Brahma, Webson, Gu, Dai, Suzgun, Chen, Chowdhery, Castro-Ros, Pellat, Robinson, Valter, Narang, Mishra, Yu, Zhao, Huang, Dai, Yu, Petrov, Chi, Dean, Devlin, Roberts, Zhou, Le, and Wei}]{chung2022scaling}
Hyung~Won Chung, Le~Hou, Shayne Longpre, Barret Zoph, Yi~Tay, William Fedus, Yunxuan Li, Xuezhi Wang, Mostafa Dehghani, Siddhartha Brahma, Albert Webson, Shixiang~Shane Gu, Zhuyun Dai, Mirac Suzgun, Xinyun Chen, Aakanksha Chowdhery, Alex Castro-Ros, Marie Pellat, Kevin Robinson, Dasha Valter, Sharan Narang, Gaurav Mishra, Adams Yu, Vincent Zhao, Yanping Huang, Andrew Dai, Hongkun Yu, Slav Petrov, Ed~H. Chi, Jeff Dean, Jacob Devlin, Adam Roberts, Denny Zhou, Quoc~V. Le, and Jason Wei. 2022.
\newblock \href {http://arxiv.org/abs/2210.11416} {Scaling instruction-finetuned language models}.

\bibitem[{Clark and Schober(1992)}]{Clark1992AskingQA}
Herbert~H. Clark and Michael~F. Schober. 1992.
\newblock \href {https://api.semanticscholar.org/CorpusID:140219575} {Asking questions and influencing answers.}

\bibitem[{DeLucia et~al.(2022)DeLucia, Wu, Mueller, Aguirre, Resnik, and Dredze}]{delucia-etal-2022-bernice}
Alexandra DeLucia, Shijie Wu, Aaron Mueller, Carlos Aguirre, Philip Resnik, and Mark Dredze. 2022.
\newblock \href {https://doi.org/10.18653/v1/2022.emnlp-main.415} {Bernice: A multilingual pre-trained encoder for {T}witter}.
\newblock In \emph{Proceedings of the 2022 Conference on Empirical Methods in Natural Language Processing}, pages 6191--6205, Abu Dhabi, United Arab Emirates. Association for Computational Linguistics.

\bibitem[{Flek(2020)}]{flek-2020-returning}
Lucie Flek. 2020.
\newblock \href {https://doi.org/10.18653/v1/2020.acl-main.700} {Returning the {N} to {NLP}: {T}owards contextually personalized classification models}.
\newblock In \emph{Proceedings of the 58th Annual Meeting of the Association for Computational Linguistics}, pages 7828--7838, Online. Association for Computational Linguistics.

\bibitem[{Fu et~al.(2020)Fu, Fussell, and Danescu-Niculescu-Mizil}]{fu-etal-2020-facilitating}
Liye Fu, Susan Fussell, and Cristian Danescu-Niculescu-Mizil. 2020.
\newblock \href {https://doi.org/10.18653/v1/2020.emnlp-main.416} {Facilitating the communication of politeness through fine-grained paraphrasing}.
\newblock In \emph{Proceedings of the 2020 Conference on Empirical Methods in Natural Language Processing (EMNLP)}, pages 5127--5140, Online. Association for Computational Linguistics.

\bibitem[{Garten et~al.(2019)Garten, Kennedy, Hoover, Sagae, and Dehghani}]{garten_incorporating_2019}
Justin Garten, Brendan Kennedy, Joe Hoover, Kenji Sagae, and Morteza Dehghani. 2019.
\newblock \href {https://doi.org/10.1111/cogs.12701} {Incorporating demographic embeddings into language understanding}.
\newblock \emph{Cognitive Science}, 43(1):e12701.

\bibitem[{Gupta et~al.(2023)Gupta, Sawant, Mishra, Nakamura, Mitra, Mashetty, and Baral}]{gupta2023instruction}
Himanshu Gupta, Saurabh~Arjun Sawant, Swaroop Mishra, Mutsumi Nakamura, Arindam Mitra, Santosh Mashetty, and Chitta Baral. 2023.
\newblock \href {http://arxiv.org/abs/2306.05539} {Instruction tuned models are quick learners}.

\bibitem[{Halliday(2004)}]{halliday2004introduction}
Michael~AK Halliday. 2004.
\newblock Introduction: How big is a language? on the power of language.
\newblock \emph{The language of science}, 5:19--32.

\bibitem[{Havaldar et~al.(2023)Havaldar, Singhal, Rai, Liu, Guntuku, and Ungar}]{havaldar-etal-2023-multilingual}
Shreya Havaldar, Bhumika Singhal, Sunny Rai, Langchen Liu, Sharath~Chandra Guntuku, and Lyle Ungar. 2023.
\newblock \href {https://doi.org/10.18653/v1/2023.wassa-1.19} {Multilingual language models are not multicultural: A case study in emotion}.
\newblock In \emph{Proceedings of the 13th Workshop on Computational Approaches to Subjectivity, Sentiment, {\&} Social Media Analysis}, pages 202--214, Toronto, Canada. Association for Computational Linguistics.

\bibitem[{Hayati et~al.(2021)Hayati, Kang, and Ungar}]{hayati-etal-2021-bert}
Shirley~Anugrah Hayati, Dongyeop Kang, and Lyle Ungar. 2021.
\newblock \href {https://doi.org/10.18653/v1/2021.emnlp-main.510} {Does {BERT} learn as humans perceive? understanding linguistic styles through lexica}.
\newblock In \emph{Proceedings of the 2021 Conference on Empirical Methods in Natural Language Processing}, pages 6323--6331, Online and Punta Cana, Dominican Republic. Association for Computational Linguistics.

\bibitem[{Heaven(2023)}]{Heaven2023ChatGPT}
William Heaven. 2023.
\newblock \href {https://www.technologyreview.com/2023/02/08/1068068/chatgpt-is-everywhere-heres-where-it-came-from/} {Chatgpt is everywhere. here’s where it came from}.
\newblock \emph{MIT Technology Review}.

\bibitem[{Hovy and Spruit(2016)}]{hovy-spruit-2016-social}
Dirk Hovy and Shannon~L. Spruit. 2016.
\newblock \href {https://doi.org/10.18653/v1/P16-2096} {The social impact of natural language processing}.
\newblock In \emph{Proceedings of the 54th Annual Meeting of the Association for Computational Linguistics (Volume 2: Short Papers)}, pages 591--598, Berlin, Germany. Association for Computational Linguistics.

\bibitem[{Hovy and Yang(2021)}]{hovy-yang-2021-importance}
Dirk Hovy and Diyi Yang. 2021.
\newblock \href {https://doi.org/10.18653/v1/2021.naacl-main.49} {The importance of modeling social factors of language: Theory and practice}.
\newblock In \emph{Proceedings of the 2021 Conference of the North American Chapter of the Association for Computational Linguistics: Human Language Technologies}, pages 588--602, Online. Association for Computational Linguistics.

\bibitem[{Hu et~al.(2022)Hu, Shen, Wallis, Allen-Zhu, Li, Wang, Wang, and Chen}]{lora}
Edward~J Hu, Yelong Shen, Phillip Wallis, Zeyuan Allen-Zhu, Yuanzhi Li, Shean Wang, Lu~Wang, and Weizhu Chen. 2022.
\newblock \href {https://openreview.net/forum?id=nZeVKeeFYf9} {Lo{RA}: Low-rank adaptation of large language models}.
\newblock In \emph{International Conference on Learning Representations}.

\bibitem[{Hu(2023)}]{Hu2023ChatGPT}
Krystal Hu. 2023.
\newblock \href {https://www.reuters.com/technology/chatgpt-sets-record-fastest-growing-user-base-analyst-note-2023-02-01/} {Chatgpt sets record for fastest-growing user base - analyst note}.
\newblock \emph{Reuters}.

\bibitem[{K{\"o}rner et~al.(2021)K{\"o}rner, Wiedemann, Hakimi, Heyer, and Potthast}]{korner-etal-2021-classifying}
Erik K{\"o}rner, Gregor Wiedemann, Ahmad~Dawar Hakimi, Gerhard Heyer, and Martin Potthast. 2021.
\newblock \href {https://doi.org/10.18653/v1/2021.emnlp-main.795} {On classifying whether two texts are on the same side of an argument}.
\newblock In \emph{Proceedings of the 2021 Conference on Empirical Methods in Natural Language Processing}, pages 10130--10138, Online and Punta Cana, Dominican Republic. Association for Computational Linguistics.

\bibitem[{Liu et~al.(2019)Liu, Ott, Goyal, Du, Joshi, Chen, Levy, Lewis, Zettlemoyer, and Stoyanov}]{liu2019roberta}
Yinhan Liu, Myle Ott, Naman Goyal, Jingfei Du, Mandar Joshi, Danqi Chen, Omer Levy, Mike Lewis, Luke Zettlemoyer, and Veselin Stoyanov. 2019.
\newblock \href {http://arxiv.org/abs/1907.11692} {Roberta: A robustly optimized bert pretraining approach}.

\bibitem[{Longpre et~al.(2023)Longpre, Hou, Vu, Webson, Chung, Tay, Zhou, Le, Zoph, Wei et~al.}]{flan-data}
Shayne Longpre, Le~Hou, Tu~Vu, Albert Webson, Hyung~Won Chung, Yi~Tay, Denny Zhou, Quoc~V Le, Barret Zoph, Jason Wei, et~al. 2023.
\newblock The flan collection: Designing data and methods for effective instruction tuning.
\newblock \emph{arXiv preprint arXiv:2301.13688}.

\bibitem[{Loshchilov and Hutter(2019)}]{adamw}
Ilya Loshchilov and Frank Hutter. 2019.
\newblock \href {http://arxiv.org/abs/1711.05101} {Decoupled weight decay regularization}.

\bibitem[{Lynn et~al.(2017)Lynn, Son, Kulkarni, Balasubramanian, and Schwartz}]{lynn-etal-2017-human}
Veronica Lynn, Youngseo Son, Vivek Kulkarni, Niranjan Balasubramanian, and H.~Andrew Schwartz. 2017.
\newblock \href {https://doi.org/10.18653/v1/D17-1119} {Human centered {NLP} with user-factor adaptation}.
\newblock In \emph{Proceedings of the 2017 Conference on Empirical Methods in Natural Language Processing}, pages 1146--1155, Copenhagen, Denmark. Association for Computational Linguistics.

\bibitem[{Matero et~al.(2019)Matero, Idnani, Son, Giorgi, Vu, Zamani, Limbachiya, Guntuku, and Schwartz}]{matero2019suicide}
Matthew Matero, Akash Idnani, Youngseo Son, Salvatore Giorgi, Huy Vu, Mohammad Zamani, Parth Limbachiya, Sharath~Chandra Guntuku, and H~Andrew Schwartz. 2019.
\newblock Suicide risk assessment with multi-level dual-context language and bert.
\newblock In \emph{Proceedings of the sixth workshop on computational linguistics and clinical psychology}, pages 39--44.

\bibitem[{Matero et~al.(2021)Matero, Soni, Balasubramanian, and Schwartz}]{matero2021melt}
Matthew Matero, Nikita Soni, Niranjan Balasubramanian, and H~Andrew Schwartz. 2021.
\newblock Melt: Message-level transformer with masked document representations as pre-training for stance detection.
\newblock \emph{arXiv preprint arXiv:2109.08113}.

\bibitem[{Meaney et~al.(2021)Meaney, Wilson, Chiruzzo, Lopez, and Magdy}]{meaney-etal-2021-semeval}
J.~A. Meaney, Steven Wilson, Luis Chiruzzo, Adam Lopez, and Walid Magdy. 2021.
\newblock \href {https://doi.org/10.18653/v1/2021.semeval-1.9} {{S}em{E}val 2021 task 7: {H}a{H}ackathon, detecting and rating humor and offense}.
\newblock In \emph{Proceedings of the 15th International Workshop on Semantic Evaluation (SemEval-2021)}, pages 105--119, Online. Association for Computational Linguistics.

\bibitem[{Mishra et~al.(2022)Mishra, Khashabi, Baral, and Hajishirzi}]{mishra-etal-2022-cross}
Swaroop Mishra, Daniel Khashabi, Chitta Baral, and Hannaneh Hajishirzi. 2022.
\newblock \href {https://doi.org/10.18653/v1/2022.acl-long.244} {Cross-task generalization via natural language crowdsourcing instructions}.
\newblock In \emph{Proceedings of the 60th Annual Meeting of the Association for Computational Linguistics (Volume 1: Long Papers)}, pages 3470--3487, Dublin, Ireland. Association for Computational Linguistics.

\bibitem[{Mohammad et~al.(2018)Mohammad, Bravo-Marquez, Salameh, and Kiritchenko}]{mohammad-etal-2018-semeval}
Saif Mohammad, Felipe Bravo-Marquez, Mohammad Salameh, and Svetlana Kiritchenko. 2018.
\newblock \href {https://doi.org/10.18653/v1/S18-1001} {{S}em{E}val-2018 task 1: Affect in tweets}.
\newblock In \emph{Proceedings of the 12th International Workshop on Semantic Evaluation}, pages 1--17, New Orleans, Louisiana. Association for Computational Linguistics.

\bibitem[{Nguyen et~al.(2020)Nguyen, Vu, and Tuan~Nguyen}]{nguyen-etal-2020-bertweet}
Dat~Quoc Nguyen, Thanh Vu, and Anh Tuan~Nguyen. 2020.
\newblock \href {https://doi.org/10.18653/v1/2020.emnlp-demos.2} {{BERT}weet: A pre-trained language model for {E}nglish tweets}.
\newblock In \emph{Proceedings of the 2020 Conference on Empirical Methods in Natural Language Processing: System Demonstrations}, pages 9--14, Online. Association for Computational Linguistics.

\bibitem[{OpenAI(2023)}]{gpt4techreport}
OpenAI. 2023.
\newblock \href {http://arxiv.org/abs/2303.08774} {Gpt-4 technical report}.

\bibitem[{Ouyang et~al.(2022)Ouyang, Wu, Jiang, Almeida, Wainwright, Mishkin, Zhang, Agarwal, Slama, Ray et~al.}]{ouyang2022training}
Long Ouyang, Jeffrey Wu, Xu~Jiang, Diogo Almeida, Carroll Wainwright, Pamela Mishkin, Chong Zhang, Sandhini Agarwal, Katarina Slama, Alex Ray, et~al. 2022.
\newblock Training language models to follow instructions with human feedback.
\newblock \emph{Advances in Neural Information Processing Systems}, 35:27730--27744.

\bibitem[{Pei and Jurgens(2020)}]{pei-jurgens-2020-quantifying}
Jiaxin Pei and David Jurgens. 2020.
\newblock \href {https://doi.org/10.18653/v1/2020.emnlp-main.428} {Quantifying intimacy in language}.
\newblock In \emph{Proceedings of the 2020 Conference on Empirical Methods in Natural Language Processing (EMNLP)}, pages 5307--5326, Online. Association for Computational Linguistics.

\bibitem[{Preo{\c{t}}iuc-Pietro et~al.(2019)Preo{\c{t}}iuc-Pietro, Gaman, and Aletras}]{preotiuc-pietro-etal-2019-automatically}
Daniel Preo{\c{t}}iuc-Pietro, Mihaela Gaman, and Nikolaos Aletras. 2019.
\newblock \href {https://doi.org/10.18653/v1/P19-1495} {Automatically identifying complaints in social media}.
\newblock In \emph{Proceedings of the 57th Annual Meeting of the Association for Computational Linguistics}, pages 5008--5019, Florence, Italy. Association for Computational Linguistics.

\bibitem[{Pryzant et~al.(2020)Pryzant, Martinez, Dass, Kurohashi, Jurafsky, and Yang}]{Subjectivebias}
Reid Pryzant, Richard~Diehl Martinez, Nathan Dass, Sadao Kurohashi, Dan Jurafsky, and Diyi Yang. 2020.
\newblock Automatically neutralizing subjective bias in text.
\newblock \emph{arXiv preprint arXiv:1911.09709}.

\bibitem[{Rajbhandari et~al.(2020)Rajbhandari, Rasley, Ruwase, and He}]{rajbhandari2020zero}
Samyam Rajbhandari, Jeff Rasley, Olatunji Ruwase, and Yuxiong He. 2020.
\newblock \href {http://arxiv.org/abs/1910.02054} {Zero: Memory optimizations toward training trillion parameter models}.

\bibitem[{Rosenthal et~al.(2017)Rosenthal, Farra, and Nakov}]{rosenthal-etal-2017-semeval}
Sara Rosenthal, Noura Farra, and Preslav Nakov. 2017.
\newblock \href {https://doi.org/10.18653/v1/S17-2088} {{S}em{E}val-2017 task 4: Sentiment analysis in {T}witter}.
\newblock In \emph{Proceedings of the 11th International Workshop on Semantic Evaluation ({S}em{E}val-2017)}, pages 502--518, Vancouver, Canada. Association for Computational Linguistics.

\bibitem[{Ruan et~al.(2016)Ruan, Wilson, and Mihalcea}]{ruan-etal-2016-finding}
Xianzhi Ruan, Steven Wilson, and Rada Mihalcea. 2016.
\newblock \href {https://doi.org/10.18653/v1/P16-2052} {Finding optimists and pessimists on {T}witter}.
\newblock In \emph{Proceedings of the 54th Annual Meeting of the Association for Computational Linguistics (Volume 2: Short Papers)}, pages 320--325, Berlin, Germany. Association for Computational Linguistics.

\bibitem[{Sanh et~al.(2022)Sanh, Webson, Raffel, Bach, Sutawika, Alyafeai, Chaffin, Stiegler, Raja, Dey, Bari, Xu, Thakker, Sharma, Szczechla, Kim, Chhablani, Nayak, Datta, Chang, Jiang, Wang, Manica, Shen, Yong, Pandey, Bawden, Wang, Neeraj, Rozen, Sharma, Santilli, Fevry, Fries, Teehan, Scao, Biderman, Gao, Wolf, and Rush}]{sanh2022multitask}
Victor Sanh, Albert Webson, Colin Raffel, Stephen Bach, Lintang Sutawika, Zaid Alyafeai, Antoine Chaffin, Arnaud Stiegler, Arun Raja, Manan Dey, M~Saiful Bari, Canwen Xu, Urmish Thakker, Shanya~Sharma Sharma, Eliza Szczechla, Taewoon Kim, Gunjan Chhablani, Nihal Nayak, Debajyoti Datta, Jonathan Chang, Mike Tian-Jian Jiang, Han Wang, Matteo Manica, Sheng Shen, Zheng~Xin Yong, Harshit Pandey, Rachel Bawden, Thomas Wang, Trishala Neeraj, Jos Rozen, Abheesht Sharma, Andrea Santilli, Thibault Fevry, Jason~Alan Fries, Ryan Teehan, Teven~Le Scao, Stella Biderman, Leo Gao, Thomas Wolf, and Alexander~M Rush. 2022.
\newblock \href {https://openreview.net/forum?id=9Vrb9D0WI4} {Multitask prompted training enables zero-shot task generalization}.
\newblock In \emph{International Conference on Learning Representations}.

\bibitem[{Sap et~al.(2020)Sap, Gabriel, Qin, Jurafsky, Smith, and Choi}]{sap-etal-2020-social}
Maarten Sap, Saadia Gabriel, Lianhui Qin, Dan Jurafsky, Noah~A. Smith, and Yejin Choi. 2020.
\newblock \href {https://doi.org/10.18653/v1/2020.acl-main.486} {Social bias frames: Reasoning about social and power implications of language}.
\newblock In \emph{Proceedings of the 58th Annual Meeting of the Association for Computational Linguistics}, pages 5477--5490, Online. Association for Computational Linguistics.

\bibitem[{Sharma et~al.(2020)Sharma, Miner, Atkins, and Althoff}]{sharma-etal-2020-computational}
Ashish Sharma, Adam Miner, David Atkins, and Tim Althoff. 2020.
\newblock \href {https://doi.org/10.18653/v1/2020.emnlp-main.425} {A computational approach to understanding empathy expressed in text-based mental health support}.
\newblock In \emph{Proceedings of the 2020 Conference on Empirical Methods in Natural Language Processing (EMNLP)}, pages 5263--5276, Online. Association for Computational Linguistics.

\bibitem[{Socher et~al.(2013)Socher, Perelygin, Wu, Chuang, Manning, Ng, and Potts}]{socher-etal-2013-recursive}
Richard Socher, Alex Perelygin, Jean Wu, Jason Chuang, Christopher~D. Manning, Andrew Ng, and Christopher Potts. 2013.
\newblock \href {https://aclanthology.org/D13-1170} {Recursive deep models for semantic compositionality over a sentiment treebank}.
\newblock In \emph{Proceedings of the 2013 Conference on Empirical Methods in Natural Language Processing}, pages 1631--1642, Seattle, Washington, USA. Association for Computational Linguistics.

\bibitem[{Touvron et~al.(2023)Touvron, Martin, Stone, Albert, Almahairi, Babaei, Bashlykov, Batra, Bhargava, Bhosale et~al.}]{touvron2023llama}
Hugo Touvron, Louis Martin, Kevin Stone, Peter Albert, Amjad Almahairi, Yasmine Babaei, Nikolay Bashlykov, Soumya Batra, Prajjwal Bhargava, Shruti Bhosale, et~al. 2023.
\newblock Llama 2: Open foundation and fine-tuned chat models.
\newblock \emph{arXiv preprint arXiv:2307.09288}.

\bibitem[{V~Ganesan et~al.(2023)V~Ganesan, Lal, Nilsson, and Schwartz}]{v-ganesan-etal-2023-systematic}
Adithya V~Ganesan, Yash~Kumar Lal, August Nilsson, and H.~Schwartz. 2023.
\newblock \href {https://doi.org/10.18653/v1/2023.wassa-1.34} {Systematic evaluation of {GPT}-3 for zero-shot personality estimation}.
\newblock In \emph{Proceedings of the 13th Workshop on Computational Approaches to Subjectivity, Sentiment, {\&} Social Media Analysis}, pages 390--400, Toronto, Canada. Association for Computational Linguistics.

\bibitem[{V~Ganesan et~al.(2021)V~Ganesan, Matero, Ravula, Vu, and Schwartz}]{v-ganesan-etal-2021-empirical}
Adithya V~Ganesan, Matthew Matero, Aravind~Reddy Ravula, Huy Vu, and H.~Andrew Schwartz. 2021.
\newblock \href {https://doi.org/10.18653/v1/2021.naacl-main.357} {Empirical evaluation of pre-trained transformers for human-level {NLP}: The role of sample size and dimensionality}.
\newblock In \emph{Proceedings of the 2021 Conference of the North American Chapter of the Association for Computational Linguistics: Human Language Technologies}, pages 4515--4532, Online. Association for Computational Linguistics.

\bibitem[{Van~Hee et~al.(2018)Van~Hee, Lefever, and Hoste}]{van-hee-etal-2018-semeval}
Cynthia Van~Hee, Els Lefever, and V{\'e}ronique Hoste. 2018.
\newblock \href {https://doi.org/10.18653/v1/S18-1005} {{S}em{E}val-2018 task 3: Irony detection in {E}nglish tweets}.
\newblock In \emph{Proceedings of the 12th International Workshop on Semantic Evaluation}, pages 39--50, New Orleans, Louisiana. Association for Computational Linguistics.

\bibitem[{Varadarajan et~al.(2022)Varadarajan, Soni, Wang, Luhmann, Schwartz, and Inoue}]{varadarajan-etal-2022-detecting}
Vasudha Varadarajan, Nikita Soni, Weixi Wang, Christian Luhmann, H.~Andrew Schwartz, and Naoya Inoue. 2022.
\newblock \href {https://doi.org/10.18653/v1/2022.nlpcss-1.16} {Detecting dissonant stance in social media: The role of topic exposure}.
\newblock In \emph{Proceedings of the Fifth Workshop on Natural Language Processing and Computational Social Science (NLP+CSS)}, pages 151--156, Abu Dhabi, UAE. Association for Computational Linguistics.

\bibitem[{Wang et~al.(2023)Wang, Kordi, Mishra, Liu, Smith, Khashabi, and Hajishirzi}]{wang-etal-2023-self-instruct}
Yizhong Wang, Yeganeh Kordi, Swaroop Mishra, Alisa Liu, Noah~A. Smith, Daniel Khashabi, and Hannaneh Hajishirzi. 2023.
\newblock \href {https://doi.org/10.18653/v1/2023.acl-long.754} {Self-instruct: Aligning language models with self-generated instructions}.
\newblock In \emph{Proceedings of the 61st Annual Meeting of the Association for Computational Linguistics (Volume 1: Long Papers)}, pages 13484--13508, Toronto, Canada. Association for Computational Linguistics.

\bibitem[{Wang et~al.(2022)Wang, Mishra, Alipoormolabashi, Kordi, Mirzaei, Naik, Ashok, Dhanasekaran, Arunkumar, Stap, Pathak, Karamanolakis, Lai, Purohit, Mondal, Anderson, Kuznia, Doshi, Pal, Patel, Moradshahi, Parmar, Purohit, Varshney, Kaza, Verma, Puri, Karia, Doshi, Sampat, Mishra, Reddy~A, Patro, Dixit, and Shen}]{wang-etal-2022-super}
Yizhong Wang, Swaroop Mishra, Pegah Alipoormolabashi, Yeganeh Kordi, Amirreza Mirzaei, Atharva Naik, Arjun Ashok, Arut~Selvan Dhanasekaran, Anjana Arunkumar, David Stap, Eshaan Pathak, Giannis Karamanolakis, Haizhi Lai, Ishan Purohit, Ishani Mondal, Jacob Anderson, Kirby Kuznia, Krima Doshi, Kuntal~Kumar Pal, Maitreya Patel, Mehrad Moradshahi, Mihir Parmar, Mirali Purohit, Neeraj Varshney, Phani~Rohitha Kaza, Pulkit Verma, Ravsehaj~Singh Puri, Rushang Karia, Savan Doshi, Shailaja~Keyur Sampat, Siddhartha Mishra, Sujan Reddy~A, Sumanta Patro, Tanay Dixit, and Xudong Shen. 2022.
\newblock \href {https://doi.org/10.18653/v1/2022.emnlp-main.340} {Super-{N}atural{I}nstructions: Generalization via declarative instructions on 1600+ {NLP} tasks}.
\newblock In \emph{Proceedings of the 2022 Conference on Empirical Methods in Natural Language Processing}, pages 5085--5109, Abu Dhabi, United Arab Emirates. Association for Computational Linguistics.

\bibitem[{Wei et~al.(2022{\natexlab{a}})Wei, Bosma, Zhao, Guu, Yu, Lester, Du, Dai, and Le}]{wei2022finetuned}
Jason Wei, Maarten Bosma, Vincent~Y. Zhao, Kelvin Guu, Adams~Wei Yu, Brian Lester, Nan Du, Andrew~M. Dai, and Quoc~V. Le. 2022{\natexlab{a}}.
\newblock \href {http://arxiv.org/abs/2109.01652} {Finetuned language models are zero-shot learners}.

\bibitem[{Wei et~al.(2022{\natexlab{b}})Wei, Tay, Bommasani, Raffel, Zoph, Borgeaud, Yogatama, Bosma, Zhou, Metzler, Chi, Hashimoto, Vinyals, Liang, Dean, and Fedus}]{wei2022emergent}
Jason Wei, Yi~Tay, Rishi Bommasani, Colin Raffel, Barret Zoph, Sebastian Borgeaud, Dani Yogatama, Maarten Bosma, Denny Zhou, Donald Metzler, Ed~H. Chi, Tatsunori Hashimoto, Oriol Vinyals, Percy Liang, Jeff Dean, and William Fedus. 2022{\natexlab{b}}.
\newblock \href {https://openreview.net/forum?id=yzkSU5zdwD} {Emergent abilities of large language models}.
\newblock \emph{Transactions on Machine Learning Research}.
\newblock Survey Certification.

\bibitem[{Wittgenstein(1953)}]{wittgenstein1953philosophical}
Ludwig Wittgenstein. 1953.
\newblock \emph{Philosophical investigations}.

\bibitem[{Wolf et~al.(2020)Wolf, Debut, Sanh, Chaumond, Delangue, Moi, Cistac, Rault, Louf, Funtowicz, Davison, Shleifer, von Platen, Ma, Jernite, Plu, Xu, Le~Scao, Gugger, Drame, Lhoest, and Rush}]{wolf-etal-2020-transformers}
Thomas Wolf, Lysandre Debut, Victor Sanh, Julien Chaumond, Clement Delangue, Anthony Moi, Pierric Cistac, Tim Rault, Remi Louf, Morgan Funtowicz, Joe Davison, Sam Shleifer, Patrick von Platen, Clara Ma, Yacine Jernite, Julien Plu, Canwen Xu, Teven Le~Scao, Sylvain Gugger, Mariama Drame, Quentin Lhoest, and Alexander Rush. 2020.
\newblock \href {https://doi.org/10.18653/v1/2020.emnlp-demos.6} {Transformers: State-of-the-art natural language processing}.
\newblock In \emph{Proceedings of the 2020 Conference on Empirical Methods in Natural Language Processing: System Demonstrations}, pages 38--45, Online. Association for Computational Linguistics.

\bibitem[{Zhang and Wan(2022)}]{zhang-wan-2022-mover}
Yunxiang Zhang and Xiaojun Wan. 2022.
\newblock \href {https://doi.org/10.18653/v1/2022.naacl-main.440} {{MOVER}: Mask, over-generate and rank for hyperbole generation}.
\newblock In \emph{Proceedings of the 2022 Conference of the North American Chapter of the Association for Computational Linguistics: Human Language Technologies}, pages 6018--6030, Seattle, United States. Association for Computational Linguistics.

\bibitem[{Ziems et~al.(2023)Ziems, Held, Shaikh, Chen, Zhang, and Yang}]{ziems2023can}
Caleb Ziems, William Held, Omar Shaikh, Jiaao Chen, Zhehao Zhang, and Diyi Yang. 2023.
\newblock \href {https://doi.org/10.1162/coli_a_00502} {{Can Large Language Models Transform Computational Social Science?}}
\newblock \emph{Computational Linguistics}, pages 1--53.

\end{thebibliography}

\clearpage

\appendix

\section{Appendix}
\label{sec:appendix}

We first provide statistics about \ourdata which has been used to instruction tune \ours.
\autoref{tab:train-data-stats} includes information about amount of data used per task as well as the number of labels used.

Next, we provide descriptions of the instructions and prompt templates used for each task in \ourdata.
Please note that this may contain examples of potentially dangerous and harmful text.

\begin{table*}
\small
\centering
\begin{tabular}{|l|r|r|r|l|}
\hline
\textbf{Dataset} & \textbf{Train Set} & \textbf{Validation Set} & \textbf{Test Set}  &\textbf{Num\_classes}\\
\hline
\emotion & 3257 & 374 & 1421  &4\\
\flute & 6780 & 754 & 1498  &4\\
\empathy & 2220 & 247 & 617  &3\\
\humcls & 8000 & 1000 & 1000  &2\\
\offense & 8000 & 4666 & 4691  &2\\
\sexyn & 7999 & 4666 & 4691  &2\\
\intent & 7999 & 4666 & 4691  &2\\
\biasedimp & 7999 & 4666 & 4691  &2\\
\politenesshayati & 256 & 32 & 32  &2\\
\hyperbole & 2580 & 323 & 323  &2\\
\sameside & 140 & 18 & 17  &2\\
\Sentiment & 8000 & 2000 & 12284  &3\\
\intimacy & 1797 & 225 & 225  &6\\
\neutralisebias & 8000 & 9379 & 9379  &2\\
\valence & 9002 & 510 & 550  &2\\
\arousal & 9002 & 510 & 550  &2\\
\dominance & 9002 & 510 & 550  &2\\
\empathyself & 1487 & 186 & 186  &2\\
\distressself & 1487 & 186 & 186  &2\\
\humrat & 4932 & 632 & 615  &2\\
\hline
\hate & - & - & 2970  &2\\
\irony & - & - & 784 &2\\
\stanfordpoliteness & - & - & 567  &2\\
\optimism & - & - & 1495 &3\\
\complaints & - & - & 345 &2\\
\kialo & - & - & 4760 &3\\
\hline
\end{tabular}
\caption{Training, validation and test set statistics \ourdata. `-' denotes that the dataset was not used in training \ours but we create instructions for its test set. Overall, \ourdata contains \textasciitilde 202k data points.}
\label{tab:train-data-stats}
\end{table*}

\begin{table*}[!t]
\small
\centering
\begin{tabular}{| l | p{1.5\columnwidth}|}
\hline
\textbf{Dataset} & \textbf{Instruction Format Example} \\
\hline
\humcls & Instruction: Upon receiving a piece of text, your task is to analyze and determine whether the intention of the text was to be humorous. You are instructed to look at the text and identify the structure of the joke, e.g. setup and punchline, or the content of the joke, e.g. absurdity, in order to determine if the intention of the text was to be humorous. If you think the intention of the text was to be humorous, classify it as `humorous', else classfy it as `non-humorous'.

Input: TENNESSEE: We're the best state. Nobody even comes close. *Elevennessee walks into the room* TENNESSEE: Oh shit...

Output: humorous\\
\hline
\humrat & Instruction: Upon receiving a piece of text, your task is to assess its comedic quality and categorize it as either `low humor' or `high humor'.

Input: How many men does it take to open a can of beer? None. It should be open by the time she brings it to the couch.

Output: low humor \\
\hline
\end{tabular}
\caption{Instruction prompts and output examples for humour task }
\label{tab:prompts-humour}
\end{table*}

\begin{table*}[!t]
\small
\centering

\begin{tabular}[width=\linewidth]{| l | 
p{1.5\columnwidth}|}
\hline
\textbf{Dataset} & \textbf{Instruction Format Example} \\
\hline
\Sentiment & Instruction: Evaluate the sentiment conveyed in the input text and determine whether it is positive, negative, or neutral. This sentiment assessment should encompass the overall sentiment of the event described within the context of the topic mentioned in the text. Your options for classification are confined to positive, negative or neutral.

Input: Few more hours to iPhone 6s launch and im still using the 4th generation 

Output: positive \\
\hline
\emotion & Instruction: Analyze the following sentence and determine the predominant emotion it displays. Your options for classification are confined to anger, joy, optimism, or sadness. Please select one emotion from the given alternatives that you believe best epitomizes the emotional context of the sentence.

Input: Worry is a down payment on a problem you may never have'. Joyce Meyer. \#motivation \#leadership \#worry 

Output: Optimism \\
\hline
\valence & Instruction: Analyze the provided text using the Valence-Arousal-Dominance model for emotional assessment. Your task is to classify the valence level it would likely elicit in an average reader, where `Low Valence' indicates a low level of pleasant feelings and `High Valence' indicates a high level of pleasant feelings. Remember, the valence scale is used to measure the degree of pleasure or displeasure a person may feel towards something. Your options for classification are confined to `Low Valence' or `High Valence'.

Input: Remember what she said in my last letter?  

Output: High Valence \\
\hline
\arousal & Instruction: Analyze the provided text using the Valence-Arousal-Dominance (VAD) emotional model. Your task is to classify the arousal level it might trigger in an average reader. Arousal, in this context, refers to the degree of energy or lethargy the text might induce. `Low Arousal' indicates a low arousal level, suggesting the text is likely to make the reader feel calm or lethargic. Conversely, `High Arousal' indicates a high arousal level, suggesting the tweet is likely to energize or excite the reader. Your options for classification are confined to `Low Arousal' or `High Arousal'.

Input: Remember what she said in my last letter? 

Output: High Arousal \\
\hline
\dominance & Instruction: Please analyze the provided text using the Valence-Arousal-Dominance (VAD) model for emotional response. Specifically, we're interested in the Dominance aspect of this model. This involves assessing the level of control or dominance the text might make an average reader feel, versus feelings of being controlled or submissive.\textbackslash{}nPlease classify this dominance level as `Low Dominance' or `High dominance'. `Low Dominance' indicates that the text is likely to evoke a low level of dominance or control in the reader, making them feel more submissive or controlled. Conversely, `High Dominance' suggests that the text would make the reader feel highly dominant or in control.\textbackslash{}nYour options for classification are confined to `Low Dominance' or `High Dominance'.

Input: Remember what she said in my last letter?

Output: High Dominance \\
\hline

\sameside & Instruction: You are provided with two pieces of text sourced from an online debate forum. Your task is to analyze and categorize these texts based on their argumentative stance. Determine whether both texts are arguing in favor of the same viewpoint or if they are opposing each other. Your options for classification are confined to `same side' or `not same side'.

Input: Legalizing gay marriage will not destroy man/woman relationships. [SEP] Why should gay marriage destroy man/woman relationships? It would just give gays the option to marry.

Output: same side \\
\hline
\end{tabular}
\caption{Instruction prompts and output examples for sentiment and emotion task }
\label{tab:prompts-sentiment-emotion}
\end{table*}

\begin{table*}
\small
\centering
\begin{tabular}{| l | p{1.5\columnwidth}|}
\hline
\textbf{Dataset} & \textbf{Instruction Format Example} \\
\hline

\neutralisebias & Instruction: Given two pieces of text, your objective is to detect subjective bias, which manifests when language that should remain neutral and impartial is influenced by feelings, opinions, or personal preferences, whether intentionally or unintentionally. If you find bias in the first sentence, indicate `first sentence' as the output; otherwise, specify `second sentence'.

Input: the term finds widespread usage among members of the educational establishment who see students as tools of social change. [SEP] the term finds widespread usage among members of the educational establishment who see students as agents of social change.

Output: first sentence \\
\hline

\hyperbole & Instruction: Upon receiving a piece of text, your task is to analyze and determine whether it contains hyperbolic language, which is an exaggerated statement or claim not meant to be taken literally, or if it does not. Your options for classification are confined to `hyperbole' or `not hyperbole'.

Input: He looks great and yet he must be pushing sixty by now.

Output: not hyperbole \\

\hline
\end{tabular}
\caption{Instruction prompts and output examples for trustworthiness task }
\label{tab:prompts-trustworthiness}
\end{table*}

\begin{table*}
\small
\centering

\begin{tabular}[width=\linewidth]{| l | p{1.5\columnwidth}|}
\hline
\textbf{Dataset} & \textbf{Instruction Format Example} \\
\hline
\empathy & Instruction: Evaluate the degree of inquiry exhibited in the counselor's response provided below, categorizing it as either ``Strong Exploration", ``Weak Exploration" or ``No Exploration". We define `exploration' as instances where a mental health counselor displays keen interest in a patient by asking about experiences that haven't been explicitly mentioned.

Input:

Patient: I'm finally on medication and found a therapist who makes me feel hopeful for the first time in years. It has been a while since I've felt sincere hope.
Counselor's response: im envious of the fact that you want hope.

Output: No Exploration \\
\hline
\empathyself & Instruction: Please carefully peruse the subsequent text, which is a personal account penned by an individual expressing their emotions and reflections after reading a news article. This account is directed towards their friends. After reading, your task is to accurately classify the level of empathetic concern demonstrated by the author. Your options for classification are `low empathy' which indicates low empathetic concern or `high empathy' which signifies a high degree of empathetic concern.

Input: This sounds like a horrible accident. I can't even imagine what that family is going through. The ones that were involved in the accident but survived probably saw some horrible things, and those kids who witnessed their mother die. As a parent myself that's a terrifying thought I would never want my son to go through something like that.

Output: high empathy \\
\hline
\distressself & Instruction: Please carefully peruse the subsequent text, which is a personal account written by an individual to their friends. This account details their emotional reactions and cognitive responses upon reading a specific news article. Your task is to accurately classify the level of personal distress experienced by the author. Your options for classification are `low distress' or `high distress'.

Input: This sounds like a horrible accident. I can't even imagine what that family is going through. The ones that were involved in the accident but survived probably saw some horrible things, and those kids who witnessed their mother die. As a parent myself that's a terrifying thought I would never want my son to go through something like that.

Output: low distress \\
\hline
\flute & Instruction: Please follow these steps:

1. First, you'll be presented with a premise and a hypothesis in the input section.

2. Your task is to determine and categorize the type of figurative language utilized in the hypothesis.

3. Finally, based on your assessment, respond with a single answer that most accurately represents the figurative language detected in the hypothesis. Choose from these four classifications: Idiom, Metaphor, Sarcasm, or Simile.

Input:

Premise: he was flying high happy and joy.

Hypothesis: he was feeling down hurt and sad.

Output: Metaphor \\
\hline

\end{tabular}
\caption{Instruction prompts and output examples for other social factors }
\label{tab:prompts-other}
\end{table*}


\begin{table*}
\small
\centering

\begin{tabular}{| l | p{1.5\columnwidth}|}
\hline
\textbf{Dataset} & \textbf{Instruction Format Example} \\ 
\hline

\politenesshayati & Instruction: Upon receiving a piece of text, your task is to analyze and determine whether the language used within it is courteous and respectful, indicating politeness, or if it contains disrespectful or rude elements, indicating impoliteness. Your options for classification are confined to `impolite' or `polite'. 

Input: @selfiequeenbri: cause I'm tired of you big bitches coming for us skinny girls!!

Output: impolite \\
\hline
\intimacy & Instruction: Assess the degree of intimacy expressed in the input text, taking into account the social context within the text. Your options for classification are confined to `very intimate`, `intimate', `somewhat intimate', `not very intimate', `not intimate' or `not intimate at all'.

Input: What are the most mediocre animals in the animal kingdom?

Output: not intimate \\
\hline
\end{tabular}
\caption{Instruction prompts and output examples for other social factors }
\label{tab:prompts-other2}
\end{table*}

\begin{table*}
\small
\centering

\begin{tabular}{| l | p{1.5\columnwidth}|}
\hline
\textbf{Dataset} & \textbf{Instruction Format Example} \\ 
\hline
\offense & Instruction: Evaluate the given text for any offensive content, which includes rudeness, disrespect, or toxicity. This assessment should consider if the text could potentially offend anyone, based on previous studies indicating a higher recall rate. Identify any disrespectful, inappropriate, or harmful language, phrases, or sentiments. If these elements exist, label the text as `offensive'. If these elements are absent, mark the text as `not offensive'.

Input: jack and jill went up a hill . <repeat> jill came down , i went up , and blew jack . ooohh ! <allcaps> it do n't rhyme . it just happened that way.

Output: not offensive \\
\hline
\sexyn & Instruction: Analyze the provided sentence and evaluate if it contains any elements that could be considered as gender-based discrimination. You are required to categorize the sentence into one of two classifications: `sexism' if it exhibits gender discrimination, or `not sexism' if it does not.

Input: "Alternative Theory: They had to change the timeline to explain why the building engineer was up there--had he heard shots, he would not have gone to 32nd floor-- in order to have \#StephenSchuck corroborate \#JesusCampos story, it had to have been before shooting started.

Output: not sexism \\
\hline
\intent & Instruction: Your task is to analyze the given text and identify if it intentionally promotes social biases and stereotypes. If you perceive that the author's motive was to deliberately offend, categorize the text as `intentional'. If there's no apparent intent to offend, classify it as `not intentional'. Remember, the focus is on the author's intent to offend, not just the offensiveness of the content.

Input: should i start discourse with eileen and hope it makes them stop telling my friends im a pedophile cus i'm fucking not?

Output: not intentional \\
\hline
\biasedimp & Instruction: Your task is to scrutinize a given text and identify any signs of prejudice. You need to differentiate between statements that merely insult an individual and those that imply group discrimination, such as comparing `F*ck you' (an individual insult) with `F*ck you, f*ggot' (a group implication). Based on your analysis, categorize the text as either `biased' or `not biased'.

Input: Obviously everyone you disagree with on twitter is a manipulation bot \& everyone you agree with is real, right? :p

Output: biased \\
\hline

\end{tabular}
\caption{Instruction prompts and output examples for offensive task }
\label{tab:prompts-offense}
\end{table*}

\begin{table*}
\small
\centering

\begin{tabular}{| l | p{12cm}|}
\hline
\textbf{Dataset} & \textbf{Instruction Format Example} \\ 
\hline

\hate & Instruction: Analyze the given text for hateful contents against women or immigrants. Your options for classification are confined to `hate speech' or `not hate speech'.

Input: Send home migrants not in need of protection, Peter Dutton tells UN, HEY DUTTON HOW ABOUT THE ONES THAT HAVE STAYED AND NOT LEFT THE COUNTRY WHEN THEY SHOULD OVERSTAYERS ? WHY DONT YOU GO AND ROUND ALL THEM UP ?

Output: hate speech \\
\hline
\irony & Instruction: Classify the input sentence as ironic or not ironic. Here ironic can be defined as the use of words or situations to convey a meaning that is opposite to what is expected. Your options for classification are confined to `ironic' or `not ironic'.

Input: Now I remember why I buy books online @user \#servicewithasmile

Output: ironic \\
\hline
\stanfordpoliteness & Instruction: Analyze the provided text, considering its tone and language, and categorize it as either polite or impolite. Your options for classification are confined to `impolite' or `polite'.

Input: Hi, I made on comment on the image page about this being different than Sbarro3 image.  Please... comment?

Output: polite \\
\hline
\optimism & Instruction: Analyze the sentiment of the provided text and classify it as `optimistic', `pessimistic' or `neutral'.

Input: fuck you fuck you fuck you

Output: pessimistic \\
\hline
\complaints & Instruction: Given an input text, identify if it contains a complaint or not. Complaining is a basic speech act used to express a negative mismatch between reality and expectations towards a state of affairs, product, organization or event. Your options for classification are confined to `complaint' or `not complaint'.

Input: @SamsungSupport Can someone please help me? I've already sent a DM.

Output: complaint \\
\hline
\kialo & Instruction: You are provided with two pieces of text and your task is to analyze and categorize these texts based on their argumentative stance. Determine whether both texts are arguing in favor of the same viewpoint, if they are opposing each other or if they are talking about two different topics. Your options for classification are confined to `agree', `disagree' or `N/A'.

Input: President Trump also intentionally lied to hide the truth from the American people and make himself look better. [SEP] President Trump told public lies every single day for his first 40 days.

Output: agree \\
\hline
\end{tabular}
\caption{Instruction prompts and output examples for related social tasks tasks }
\label{tab:prompts-unseen}
\end{table*}

\end{document}